\documentclass[sn-mathphys]{sn-jnl}

\usepackage{times}
\usepackage{epsfig}
\usepackage{graphicx}
\usepackage{amsmath}
\usepackage{amssymb} 
\usepackage{bm}
\usepackage{booktabs}
\usepackage{pifont}
\usepackage{colortbl}
\usepackage{adjustbox}
\usepackage[style=base]{caption}
\usepackage[accsupp]{axessibility} 

\jyear{2023}%

\theoremstyle{thmstyleone}%
\theoremstyle{thmstyletwo}%

\theoremstyle{thmstylethree}%

\begin{document}

\title[X-Align++]{X-Align++: Cross-Modal Cross-View Alignment for Bird's-Eye-View Segmentation}


\author*[1]{\fnm{Shubhankar} \sur{Borse}$^{\dag}$}\email{sborse@qti.qualcomm.com}

\author[2]{\fnm{Senthil} \sur{Yogamani}$^{\dag}$}\email{syogaman@qti.qualcomm.com}

\author[3]{\fnm{Marvin} \sur{Klingner}}\email{mklingne@qti.qualcomm.com}

\author[3]{\fnm{Varun} \sur{Ravi}}\email{vravikum@qti.qualcomm.com}

\author[1]{\fnm{Hong} \sur{Cai}}\email{hongcai@qti.qualcomm.com}

\author[4]{\fnm{Abdulaziz} \sur{Almuzairee}}\email{aalmuzairee@ucsd.edu}

\author[1]{\fnm{Fatih} \sur{Porikli}}\email{fporikli@qti.qualcomm.com}

\affil[1]{\orgdiv{Qualcomm AI Research}, \orgname{an initiative of Qualcomm Technologies, Inc}}

\affil[2]{\orgdiv{Automated Driving}, \orgname{QT Technologies Ireland Limited}}

\affil[3]{\orgdiv{Automated Driving}, \orgname{Qualcomm Technologies, Inc}}

\affil[4]{\orgdiv{University of California}, \orgname{San Diego}} 

\affil[\textbf{\dag}]{\orgdiv{co-first authors}}


\definecolor{darkgreen}{RGB}{0, 150, 0}
\definecolor{darkred}{RGB}{200, 0, 0}
\definecolor{darkblue}{RGB}{0, 0, 200}
\newcommand{\ch}{{\color{darkgreen} \ding{51}}}
\newcommand{\xm}{{\color{darkred} \ding{55}}}

\definecolor{gray9}{gray}{.9}
\definecolor{gray95}{gray}{.95}
\definecolor{gray8}{gray}{.8}
\definecolor{gray85}{gray}{.85}

\abstract{Bird's-eye-view (BEV) grid is a typical representation of the perception of road components, e.g., drivable area, in autonomous driving. Most existing approaches rely on cameras only to perform segmentation in BEV space, which is fundamentally constrained by the absence of reliable depth information. The latest works leverage both camera and LiDAR modalities but suboptimally fuse their features using simple, concatenation-based mechanisms. In this paper, we address these problems by enhancing the alignment of the unimodal features in order to aid feature fusion, as well as enhancing the alignment between the cameras' perspective view (PV) and BEV representations. We propose X-Align, a novel end-to-end cross-modal and cross-view learning framework for BEV segmentation consisting of the following components: (i) a novel Cross-Modal Feature Alignment (X-FA) loss, (ii) an attention-based Cross-Modal Feature Fusion (X-FF) module to align multi-modal BEV features implicitly, and (iii) an auxiliary PV segmentation branch with Cross-View Segmentation Alignment (X-SA) losses to improve the PV-to-BEV transformation. We evaluate our proposed method across two commonly used benchmark datasets, i.e., nuScenes and KITTI-360. Notably, X-Align significantly outperforms the state-of-the-art by 3 absolute mIoU points on nuScenes. We also provide extensive ablation studies to demonstrate the effectiveness of the individual components.}

\maketitle 
\section{Introduction}
\begin{figure}[h]
    \centering
    \includegraphics[width=0.9\linewidth]{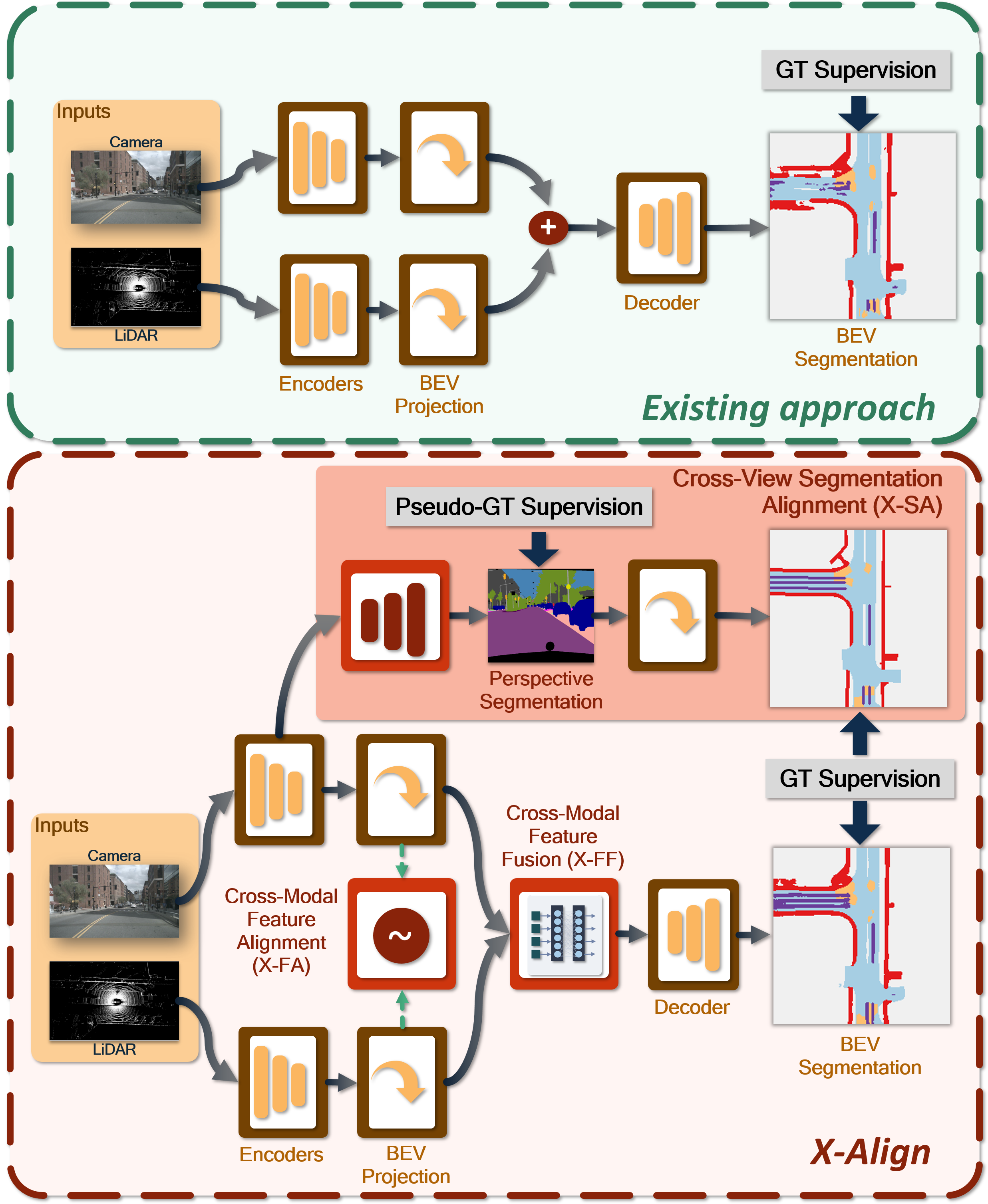}
    \caption{Existing methods for cross-modal BEV segmentation (top) utilize simple concatenation-based fusion (e.g., \cite{liu2022bevfusion}), while our proposed \textbf{X-Align enforces cross-modal feature alignment together with attention-based feature fusion, as well as cross-view segmentation alignment} (bottom). These contributions improve both feature aggregation and PV-to-BEV transformation, leading to more accurate BEV segmentation.}
    \label{fig:contribution}
\end{figure}
Bird's-eye-view (BEV) segmentation aims at classifying each cell in a BEV grid around the ego position. As such, it is a key enabler for applications like autonomous driving and robotics. For instance, the BEV segmentation map is a prerequisite for current works on behavior prediction and trajectory planning~\cite{hu2021fiery, tacs2016functional, wu2020motionnet}. It is also an important input modality for learning end-to-end controls (e.g., speed control, steering) in autonomous driving~\cite{chitta2021neat}.\par

Given the ubiquity of camera sensors, existing BEV segmentation methods predominantly focus on predicting BEV segmentation maps from camera images \cite{philion2020lift, xie2022m2bev, xu2022cobevt, zhou2022crossview}. However, the lack of reliable 3D information significantly limits the performance of these methods. A possible way to resolve this challenge is to leverage a LiDAR sensor and fuse the measured sparse geometric information with that contained in camera images. While camera-LiDAR fusion has been extensively studied for the task of 3D object detection, such fusion strategies are relatively unexplored for BEV segmentation. The concurrent work of~\cite{liu2022bevfusion} provides the first baseline using a simple concatenation of LiDAR BEV features and camera features projected from perspective view (PV) to BEV via estimated depth and voxel pooling. However, the PV-to-BEV projection can be inaccurate due to errors in depth estimation. As a result, in the concatenation stage, the network may aggregate poorly aligned features across the camera and the LiDAR branches, resulting in suboptimal fusion results.\par

In this paper, we propose a novel \textit{cross-modal, cross-view alignment strategy, X-Align}, which enforces feature alignment across features extracted from the camera and LiDAR inputs as well as segmentation consistency across PV and BEV to improve the overall BEV segmentation accuracy (cf.~Fig.~\ref{fig:contribution}). First, we propose a Cross-Modal Feature Alignment (X-FA) loss function that promotes the correlation between projected camera features and LiDAR features, as measured by cosine similarity. In addition, we incorporate attention to the Cross-Modal Feature Fusion (X-FF) of these two sets of modality-specific features instead of using simple concatenation as in~\cite{liu2022bevfusion}. This gives the network a more substantial capability to properly align and aggregate features from the two sensing modalities.\par

We further impose Cross-View Segmentation Alignment (X-SA) losses during training. We introduce a trainable segmentation decoder based on the intermediate PV camera features to generate PV segmentation. Next, we utilize the same PV-to-BEV transformation~\cite{philion2020lift} that converts PV camera features to BEV to convert the PV segmentation map into a BEV segmentation map, which is then supervised by the ground truth. We also supervise the intermediate PV segmentation map using pseudo-labels generated by a high-quality, off-the-shelf semantic segmentation model. In this way, the camera branch learns to derive intermediate features containing useful PV semantic features, providing richer information for BEV segmentation after being projected to BEV space. Moreover, this provides additional supervision on the PV-to-BEV module, allowing it to learn a more accurate transformation.\par

Our main contributions are summarized as follows:
\begin{itemize}
\setlength\itemsep{0pt}
    \item We propose a novel framework, X-Align, that enables better feature alignment and fusion across camera and LiDAR modalities and enforces segmentation alignment across perspective view and bird's eye view.  
    \item Specifically, we propose a Cross-Modal Feature Alignment (X-FA) loss to enhance the correlation between the camera and LiDAR features. We also devise an attention-based Cross-Modal Feature Fusion (X-FF).
    \item We further propose to enforce Cross-View Segmentation Alignment (X-SA) across the perspective view and bird's eye view, which encourages the model to learn richer semantic features and a more accurate PV-to-BEV projection.
    \item We conduct extensive experiments on the nuScenes and KITTI-360 datasets with comprehensive ablation studies that demonstrate the efficacy of X-Align. In particular, on nuScenes, we surpass the state-of-the-art in BEV segmentation by 3 absolute mIoU points.
\end{itemize}
\section{Related Work}
 \vspace{-2pt}

\textbf{BEV Segmentation}:
The task of BEV segmentation has mostly been explored using (multiple) camera images as input. Building on top of Perspective View (PV) segmentation~\cite{borse2021inverseform, borse2021hs3, borse2022panoptic, zhang2022auxadapt, borse2023dejavu, hu2022learning, zhang2022perceptual}, early works used the homography transformation to convert camera images to BEV, subsequently estimating the segmentation map~\cite{abbas2019geometric, garnett2019lanenet, loukkal2021driving, zhu2021monocular}. As the homography transformation introduces strong artifacts, subsequent works moved towards depth estimation and voxelization~\cite{philion2020lift, roddick2018orthographic} for the PV-to-BEV transformation as end-to-end learning~\cite{lu2019monocular, roddick2020predicting}. This basic setup has been further explored in various directions: VPN~\cite{pan2020cross} explores domain adaptation, BEVerse~\cite{zhang2022beverse} and M\textsuperscript{2}BEV~\cite{xie2022m2bev} explore multi-task learning with 3D object detection, CoBEVT~\cite{xu2022cobevt} explores fusion of features from vehicles, Gosala \textit{et al}. explore panoptic BEV segmentation~\cite{gosala22panopticbev}, while several works explore incorporation of temporal context~\cite{hu2021fiery, saha2021enabling}. Furthermore, CVT~\cite{zhou2022crossview} uses a learned map embedding and an attention mechanism between map queries and camera features. In contrast to these existing BEV segmentation approaches that only use camera images, we explore the multi-modal fusion of LiDAR point clouds and camera images.\par

Multi-modal fusion for BEV segmentation has been enabled by recently introduced large-scale datasets providing time-synchronized data from multiple sensors~\cite{caesar2020nuscenes, geiger2013vision, sun2020waymo}. However, most works on these datasets focus on the 3D object detection task~\cite{klingner2023x, li2022deepfusion, liu2022bevfusion, yin2021center, chitta2022transfuser, shao2022safety}, while we focus on BEV segmentation. The closest prior art to our work is BEVFusion~\cite{liu2022bevfusion}. While their method also predicts BEV segmentation based on LiDAR point clouds and camera images, they use a simple feature concatenation to fuse multi-modal features such that the network implicitly has to connect information from misaligned features. In contrast, we explicitly enforce alignment between multi-modal features. Also, we enforce alignment between PV and BEV segmentation to improve the PV-to-BEV transformation.\par

\textbf{Camera-LiDAR Sensor Fusion}:
The vast majority of fusion methods have been proposed for the 3D object detection task. Initially, two-stage approaches have been proposed, lifting image bounding box proposals into 3D frustum view~\cite{nabati2021centerfusion, qi2018frustrum, wang2019frustum} for fusion with LiDAR. However, research focus has shifted towards end-to-end training, where approaches can roughly be divided into three categories: point-/input-level decoration, feature-level fusion, and proposal-level fusion. Point-level fusion includes methods such as PointAugmenting~\cite{wang2021pointaugmenting}, PointPainting~\cite{vora2020pointpainting}, FusionPainting~\cite{xu2021fusionpainting}, AutoAlign~\cite{chen2022autoalign}, and MVP~\cite{yin2021multimodal}, which extract camera features and use these features to enrich the point-level information, which is subsequently processed by a LiDAR-based detector. Recently proposed FocalSparseCNNs~\cite{chen2022focal} similarly enriches the features in the early feature extraction stage. For proposal-level fusion usually the predicted bounding boxes are refined~\cite{liang2022bevfusion}, often making use of an attention mechanism such as in FUTR3D~\cite{chen2022futr3d} and TransFusion~\cite{bai2022transfusion}. However, these two fusion types have downsides regarding generalization. While proposal-level fusion is not easily generalizable to other tasks, input-level decoration is not generically extendable to other sensor modalities.\par

Feature-level fusion aims at fusing extracted features from different sensors, subsequently predicting outputs for one or several tasks~\cite{li2022deepfusion, li2022bevformer, liang2018deep, liu2022bevfusion}. Our approach also falls into this category. While \cite{liang2018deep, liu2022bevfusion} use fusion via concatenation, more recent approaches apply attention-based fusion~\cite{li2022deepfusion, li2022bevformer}. However, these approaches still try to implicitly learn the interconnection between cross-modal features, while we explicitly encourage alignment between features from different modalities. Also, none of the above-mentioned methods uses attention-based cross-modal fusion for BEV segmentation or output-level segmentation alignment of PV and BEV segmentation.\par
\section{Proposed X-Align Framework}
\vspace{-2pt}
\begin{figure*}
    \centering
    \includegraphics[width=0.98\textwidth]{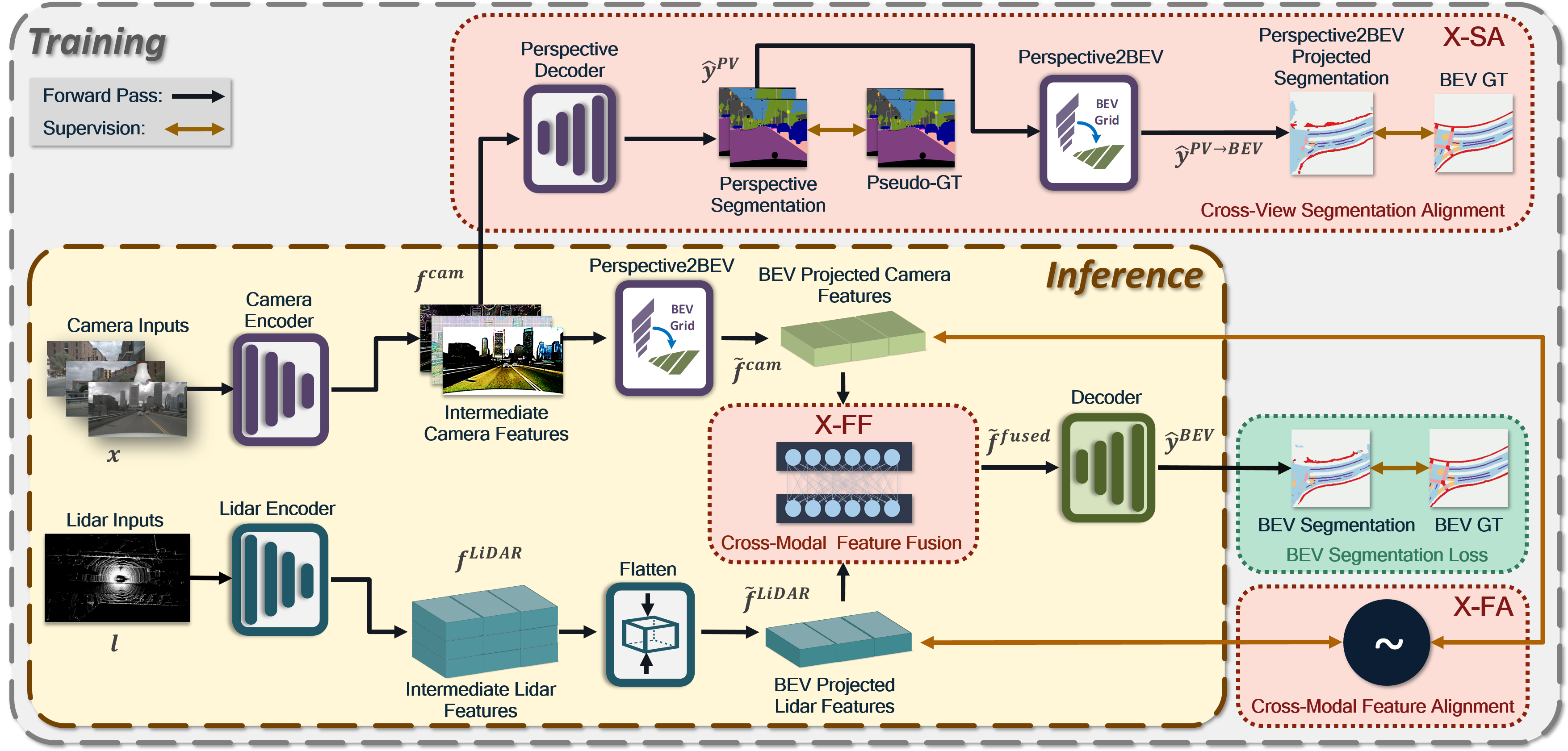}
    \caption{\textbf{Our proposed X-Align framework}: We present a cross-modal and cross-view alignment algorithm for the task of BEV segmentation based on camera images and LiDAR point clouds. We apply the Cross-View Segmentation Alignment (X-SA) and Cross-Modal Feature Alignment (X-FA) losses during training. We also propose a Cross-Modal Feature Fusion (X-FF) module to correct pixel inconsistencies between multi-modal features. Our proposed contributions are highlighted in \textcolor[HTML]{ff6865}{red}. During inference, we can remove the blocks which solely contribute to computing loss functions, suggesting that the performance enhancement comes with no added inference cost.} 
    \label{fig:method_overview}
\end{figure*}

This section describes X-Align, our novel cross-modal and cross-view alignment strategy. We first formally introduce the problem and describe a baseline method in Section~\ref{sec:method_formulation}. We provide an overview of X-Align in Section~\ref{sec:method_overview} and then discuss its components in detail, including Cross-Modal Feature Fusion (X-FF), Cross-Modal Feature Alignment (X-FA), and Cross-View Segmentation Alignment (X-SA) in Sections~\ref{sec:feature_fusion}, \ref{sec:feature_alignment}, \ref{sec:segmentation_alignment}, respectively. \par

\subsection{Problem Formulation and Baseline}\label{sec:method_formulation}
\vspace{-4pt}

Our goal is to develop a framework taking multi-modal sensor data $\mathcal{X}$ as input and predicting a BEV segmentation map $\hat{\bm{m}}\in\mathcal{S}^{H^{\mathrm{BEV}}\times W^{\mathrm{BEV}}}$, with resolution $H^{\mathrm{BEV}}\times W^{\mathrm{BEV}}$, and the set of classes $\mathcal{S}=\left\lbrace0, 1, \dots, \|\mathcal{S}\| \right\rbrace$. As illustrated in Fig.~\ref{fig:method_overview}, the set of inputs, $\mathcal{X} = \left\lbrace\bm{x}, \bm{l}\right\rbrace$, contains RGB camera images in PV, $\bm{x}\in\mathbb{R}^{N^{\mathrm{cam}}\times H^{\mathrm{cam}}\times W^{\mathrm{cam}}\times 3}$, where $N^{\mathrm{cam}}$, $H^{\mathrm{cam}}$, $W^{\mathrm{cam}}$ denote number of cameras, image height, and image width, respectively, as well as a LiDAR point cloud, $\bm{l}\in\mathbb{R}^{P\times 5}$, with number of points $P$. Each point consists of its 3-dimensional coordinates, reflectivity, and ring index. 
\par 
\textbf{Baseline Method}:
We first establish a baseline method of fusion-based BEV segmentation based on BEVFusion~\cite{liu2022bevfusion}.
As shown in Fig.~\ref{fig:method_overview}, initial features are extracted from both sensor inputs. For camera images, a camera encoder, $\bm{E}^{\mathrm{cam}}$, extracts features in PV, $\bm{f}^{\mathrm{cam}}$. Subsequently, we use a feature pyramid network (FPN) and a PV-to-BEV transformation based on~\cite{philion2020lift} to obtain camera features in BEV space, following BEVDet~\cite{huang2021bevdet}. For the LiDAR points, we follow SECOND~\cite{yan2018second} in using voxelization and a sparse LiDAR encoder, $\bm{E}^{\mathrm{LiDAR}}$. The LiDAR features are projected to BEV space using a flattening operation as in~\cite{liu2022bevfusion}. 

These operations result in two sets of modality-specific BEV features, $\bm{\Tilde{f}}^{\mathrm{cam}}\in \mathbb{R}^{H^{\mathrm{lat}}\times W^{\mathrm{lat}}\times C^{\mathrm{cam}}}$ and $\bm{\Tilde{f}}^{\mathrm{LiDAR}}\in \mathbb{R}^{H^{\mathrm{lat}}\times W^{\mathrm{lat}}\times C^{\mathrm{LiDAR}}}$, with BEV space feature resolution $H^{\mathrm{lat}}\times W^{\mathrm{lat}}$ and number of channels $C^{\mathrm{cam}}$ and $C^{\mathrm{LiDAR}}$ for camera and LiDAR features, respectively. The features are then combined (e.g., by simple concatenation as in~\cite{li2021hdmapnet, liu2022bevfusion, harley2022simple}), resulting in fused features $\bm{\Tilde{f}}^{\mathrm{fused}}\in \mathbb{R}^{H^{\mathrm{lat}}\times W^{\mathrm{lat}}\times C^{\mathrm{fused}}}$, which are further processed by a BEV Encoder and FPN as in SECOND~\cite{yan2018second}. Finally, the features are processed by a segmentation head with the same architecture as in~\cite{liu2022bevfusion} to ensure comparability. This baseline model is trained using the focal cross-entropy loss~\cite{lin2017focal}:
\begin{equation}
    \mathcal{L}^{\mathrm{BEV}} = \mathrm{FocalCE}\left (\hat{\bm{y}}^{\mathrm{BEV}}, \bm{y}^{\mathrm{BEV}}\right), 
    \label{eq:focal_bev}
\end{equation}
where $\hat{\bm{y}}^{\mathrm{BEV}}\in\mathbb{R}^{H^{\mathrm{BEV}}\times W^{\mathrm{BEV}}\times S}$ are class probabilities and $\bm{y}^{\mathrm{BEV}}\in \left\lbrace 0,1\right\rbrace^{H^{\mathrm{BEV}}\times W^{\mathrm{BEV}}\times S}$ denotes the one-hot encoded ground truth. We can obtain the final classes $\hat{\bm{m}}$ by a pixel-wise argmax operation on $\hat{\bm{y}}$ during inference.\par

\subsection{X-Align Overview}\label{sec:method_overview}
\vspace{-4pt}

Building on top of the previously described baseline, we present our novel cross-modal and cross-view alignment strategy, X-Align (highlighted by red boxes in Fig.~\ref{fig:method_overview}). First, we improve the simple concatenation-based fusion with a Cross-Modal Feature Fusion (X-FF) module, which leverages attention and mitigates misalignment between features across modalities (Section~\ref{sec:feature_fusion}). Secondly, we propose a Cross-Modal Feature Alignment (X-FA) loss, $\mathcal{L}^{\mathrm{X\text{-}FA}}$, which promotes the correlation between features across modalities (Section~\ref{sec:feature_alignment}). Finally, we propose losses to enforce Cross-View Segmentation Alignment (X-SA) in Section~\ref{sec:segmentation_alignment}, where we supervise a PV segmentation predicted from the intermediate camera features with a loss $\mathcal{L}^{\mathrm{PV}}$ and the PV-to-BEV-projected version of this segmentation with a loss $\mathcal{L}^{\mathrm{PV2BEV}}$. These two losses provide more direct training signals to the PV-to-BEV transformation and encourage richer semantic features in PV before the transformation. Overall, our total optimization objective is
\begin{equation}
    \mathcal{L}^{\mathrm{X\text{-}Align}}\! =\! \lambda_1 \mathcal{L}^{\mathrm{BEV}}\! +\! \lambda_2 \mathcal{L}^{\mathrm{X\text{-}FA}}\! +\! \lambda_3 \mathcal{L}^{\mathrm{PV}}\! +\! \lambda_4 \mathcal{L}^{\mathrm{PV2BEV}},
\label{eq:final_loss}
\end{equation}
where $\lambda_{i},\, i\in\left\lbrace 1,2,3,4 \right\rbrace$ are the loss weighting factors.

\subsection{Cross-Modal Feature Fusion (X-FF)}
\label{sec:feature_fusion}
\begin{figure}[h]
    \centering    \includegraphics[width=0.95\textwidth]{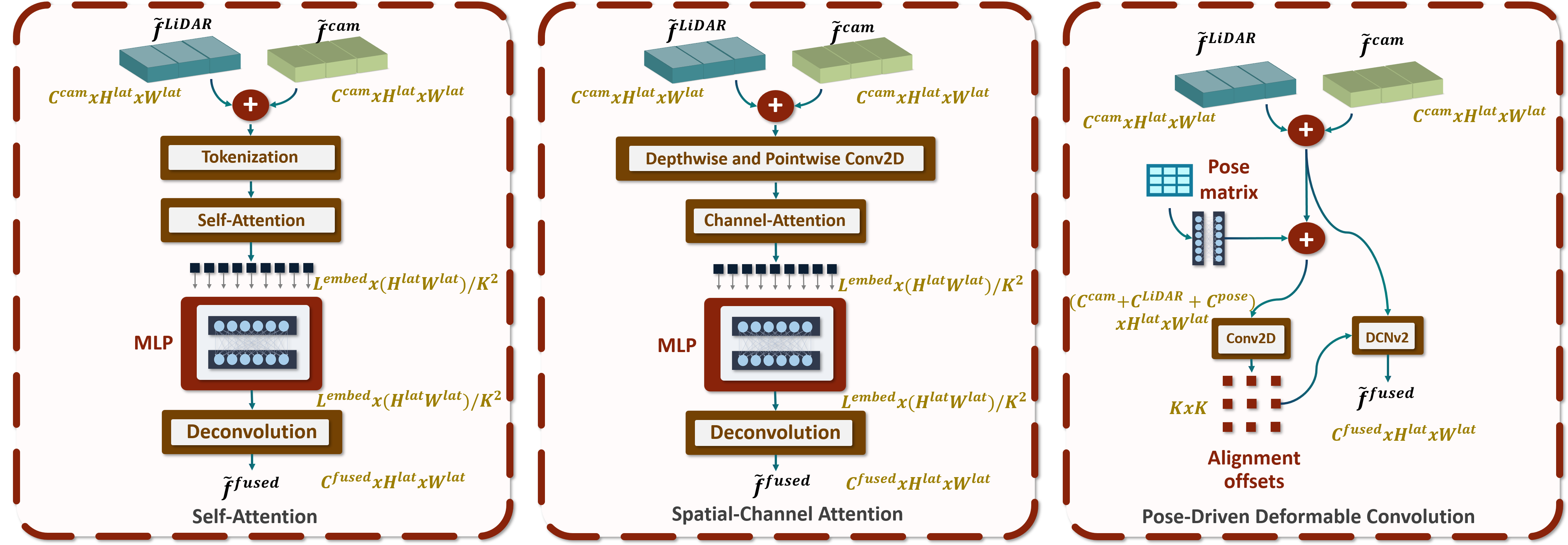}
    \caption{Our three \textbf{proposed Cross-Modal Feature Fusion (X-FF) strategies}, using standard self-attention (left), spatial-channel attention (middle), and pose-driven deformable convolution (right), respectively.}
    \label{fig:method_detail1}
\end{figure}

\vspace{-4pt}
In recent BEV segmentation literature~\cite{liu2022bevfusion, li2021hdmapnet, harley2022simple}, it is a common approach to utilize concatenation followed by convolution to combine features from multi-modal inputs, $\bm{\tilde{f}}^{\mathrm{cam}}$ and $\bm{\tilde{f}}^{\mathrm{LiDAR}}$, resulting in the aggregated features $\bm{\tilde{f}}^{\mathrm{fused}}$. However, the lack of reliable depth information can cause inaccurate PV-to-BEV transformation of features, which subsequently results in suboptimal alignment and fusion of multi-modal features. The convolution blocks utilized in existing approaches~\cite{liu2022bevfusion, li2021hdmapnet, harley2022simple} cannot rectify such misalignment due to their translation invariance. To address this issue, we propose more powerful, Cross-Modal Feature Fusion (X-FF) modules that can account for pixel-wise misalignment, as shown in Fig.~\ref{fig:method_detail1}. Next, we describe in detail our three proposed fusion designs.\par

\textbf{Self-Attention:} Our proposed X-FF using self-attention is shown in Fig.~\ref{fig:method_detail1} (left). We first stack features $\bm{\Tilde{f}}^{\mathrm{cam}}\in \mathbb{R}^{H^{\mathrm{lat}}\times W^{\mathrm{lat}}\times C^{\mathrm{cam}}}$ and $\bm{\Tilde{f}}^{\mathrm{LiDAR}}\in \mathbb{R}^{H^{\mathrm{lat}}\times W^{\mathrm{lat}}\times C^{\mathrm{LiDAR}}}$, and tokenize them into $K\times K$ patches with an embedding dimension of $L^{\mathrm{embed}}$. These patches are fed into a multi-head self-attention module~\cite{vaswani2017attention}. The output is then projected back to the original resolution using a deconvolution block, resulting in the final fused features $\bm{\Tilde{f}}^{\mathrm{fused}}\in \mathbb{R}^{H^{\mathrm{lat}}\times W^{\mathrm{lat}}\times C^{\mathrm{fused}}}$. By using self-attention, our proposed module can correspond to the camera and LiDAR features spatially, accounting for potential misalignment.\par
\textbf{Spatial-Channel Attention:} In this option, we leverage the recently proposed Split-Depth Transpose Attention (SDTA)~\cite{maaz2022edgenext}, as shown in Fig.~\ref{fig:method_detail1} (middle). It first performs spatial and channel mixing of the stacked camera and LiDAR features via depth-wise and point-wise convolutions. Then, it applies channel attention followed by a lightweight MLP. The output is passed through a deconvolution block to generate the fused features $\bm{\Tilde{f}}^{\mathrm{fused}}$. Spatial and channel mixing together with channel attention provides powerful capacity for the module to better address the misalignment between the camera and LiDAR features.\par 
\textbf{Pose-Driven Deformable Convolution:} This design is illustrated in Fig.~\ref{fig:method_detail1} (right). We know that the transformation between modalities is a function of their relative poses to the ego vehicle. Hence, we apply an adaptive transformation, i.e., Deformable Convolution (DCNv2)~\cite{zhu2019deformable}, to the stacked multi-modal features $\bm{\Tilde{f}}^{\mathrm{cam}}$ and $\bm{\Tilde{f}}^{\mathrm{LiDAR}}$, which can implicitly learn the cross-modal alignment based on available pose information. More specifically, we process the pose matrices with an MLP to generate a pose embedding $\bm{\Tilde{f}}^{\mathrm{pose}}\in \mathbb{R}^{H^{\mathrm{lat}}\times W^{\mathrm{lat}}\times C^{\mathrm{pose}}}$, which is then concatenated with $\bm{\Tilde{f}}^{\mathrm{cam}}$ and $\bm{\Tilde{f}}^{\mathrm{LiDAR}}$. They are used to generate $K\times K$ offset vectors to be used by the DCNv2 block, which produces the output fused features $\bm{\Tilde{f}}^{\mathrm{fused}}$. \par

Our proposed X-FF designs provide the network with the suitable capacity to properly align and fuse multi-modal features. While they introduce additional computations, they show more superior accuracy-efficiency trade-offs as compared to naively increasing the complexity of the baseline network, as we shall see in our study in Section~\ref{sec:ablation}.
\subsection{Cross-Modal Feature Alignment (X-FA)}
\label{sec:feature_alignment}
\vspace{-4pt}

While our proposed X-FF modules can improve feature alignment, they incur additional computations, which may not always be feasible. As such, we propose a second measure to improve feature alignment with a Cross-Modal Feature Alignment (X-FA) loss $\mathcal{L}^{\mathrm{X\text{-}FA}}$ that is only applied during training and does not introduce additional computations for inference. It can also be used in conjunction with X-FF.\par 

Consider extracted features in BEV space, $\bm{\Tilde{f}}^{\mathrm{cam}}$ and $\bm{\Tilde{f}}^{\mathrm{LiDAR}}$, from camera and LiDAR branches, respectively. We promote the correlation between the two sets of features by imposing a cosine similarity loss between them:
\begin{equation}
    \mathcal{L}^{\mathrm{X\text{-}FA}} = \mathrm{CosineSim}\left (\bm{\Tilde{f}}^{\mathrm{cam}}, \bm{\Tilde{f}}^{\mathrm{LiDAR}}\right). 
    \label{eq:feature_alignment}
\end{equation}

In order to apply this loss, there are two requirements. First, the camera features $\bm{\Tilde{f}}^{\mathrm{cam}}$ and LiDAR features $\bm{\Tilde{f}}^{\mathrm{LiDAR}}$ in BEV space need to have the same resolution. In this work, we ensure that the network parameters are chosen accordingly. In case both features have different resolutions, differentiable grid sampling~\cite{jaderberg2015spatial} can be used. Second, the channels $C^{\mathrm{cam}}$ and $C^{\mathrm{LiDAR}}$ should match for Eq.~(\ref{eq:feature_alignment}) to be applicable. This, however, is generally not the case. As such, we take all features from the lower-dimensional branch and enforce their similarity to several subsets of features from the higher-dimensional branch.

\subsection{Cross-View Segmentation Alignment (X-SA)}
\label{sec:segmentation_alignment}
\vspace{-4pt}

In addition to encouraging feature alignment, we also impose alignment at the output segmentation level across PV and BEV using our Cross-Modal Segmentation Alignment (X-SA) losses. More specifically, for the camera branch, we take intermediate PV features and feed them through an additional decoder to generate a PV segmentation prediction $\hat{\bm{y}}^{\mathrm{PV}}\in\mathbb{R}^{N^{\mathrm{cam}}\times H^{\mathrm{cam}}\times W^{\mathrm{cam}}\times\|\mathcal{S}\|}$ (cf.~Fig.~\ref{fig:method_overview}, top part). We further transform $\hat{\bm{y}}^{\mathrm{PV}}$ to BEV space by utilizing the PV-to-BEV transformation as for the features, resulting in a projected BEV segmentation map $\hat{\bm{y}}^{\mathrm{PV} \rightarrow \mathrm{BEV}}\in\mathbb{R}^{H^{\mathrm{BEV}}\times W^{\mathrm{BEV}}\times \|\mathcal{S}\|}$. The projected BEV segmentation map is supervised w.r.t ground-truth BEV segmentation using a focal cross-entropy loss:
\begin{equation}
    \mathcal{L}^{\mathrm{PV2BEV}} = \mathrm{FocalCE}\left (\hat{\bm{y}}^{\mathrm{PV} \rightarrow \mathrm{BEV}}, \bm{y}^{\mathrm{BEV}}\right). 
\end{equation}

As for the PV segmentation, since PV ground truth is not always available on BEV perception datasets, we supervise it with a focal loss using pseudo-labels generated by a state-of-the-art model pretrained on Cityscapes~\cite{cordts2016cityscapes}, as follows:
\begin{equation}
    \mathcal{L}^{\mathrm{PV}} = \mathrm{FocalCE}\left (\hat{\bm{y}}^{\mathrm{PV}}, \bm{y}^{\mathrm{PV}}\right), 
\end{equation}

By introducing these two additional supervisions, we enforce that across PV and BEV, the segmentations are accurate and aligned through the PV-to-BEV transformation. This benefit is two-fold: First, the module used here is given by the same PV-to-BEV transformation as on the feature level in the main camera branch. Our X-SA loss provides additional supervision to more accurately train this key module. Second, imposing a PV segmentation loss $\mathcal{L}^{\mathrm{PV}}$ encourages the network to learn useful PV semantic features, providing richer semantic information for the downstream BEV features. Our X-SA components, including the additional decoder and the losses, are only used during training and do not require overhead at test time.\par

In summary, our complete X-Align framework \textbf{X-Align}$_{\mathrm{all}}$
proposes four additions to the baseline: the X-FF feature fusion module, along with three additional training losses: $\mathcal{L}^{\mathrm{X\text{-}FA}}$, $\mathcal{L}^{\mathrm{PV}}$, and $\mathcal{L}^{\mathrm{PV2BEV}}$. In cases where the network only takes camera inputs, we apply the two X-SA losses, $\mathcal{L}^{\mathrm{PV}}$ and $\mathcal{L}^{\mathrm{PV2BEV}}$, giving us the \textbf{X-Align}$_{\mathrm{view}}$ variant. In addition, in case extra computation is not allowed, we apply all three X-Align losses when training the network, forming the \textbf{X-Align}$_{\mathrm{losses}}$ variant. We extensively evaluate these variants as well as combinations of our proposed X-Align components in Section~\ref{sec:ablation}.

\section{Experiments}
\label{sec:experiments}

In this section, we present comprehensive performance evaluations of X-Align and compare it with baselines and the current state of the art. We further evaluate the benefits in performance v/s added complexity of the three X-FF blocks and finally show qualitative results.\par
\begin{table}[h]
\fontsize{6.5pt}{7.5pt}\selectfont
\centering
\setlength{\tabcolsep}{2.4pt}
    \centering
    \begin{tabular}{l|c|c|cccccc|c}
    \toprule
    \textit{Model} & 
    \textit{Backbone} & 
    \textit{Modality} & 
    \cellcolor[HTML]{96bbce}\textit{Drivable} & 
    \cellcolor[HTML]{fb9a99}\textit{Ped. Cross.} & 
     \cellcolor[HTML]{fc4538}\textit{Walkway} & 
    \cellcolor[HTML]{fdbf6f}\textit{Stop} 
    \textit{Line} & 
    \cellcolor[HTML]{ff7f00}\textit{Carpark} & 
    \cellcolor[HTML]{ab9ac0}\textit{Divider} & 
    \cellcolor[HTML]{a5eb8d}\textit{mIoU} \\
    \midrule
    OFT~\cite{roddick2018orthographic} & ResNet-18 & C & 74.0 & 35.3 & 45.9 & 27.5 & 35.9 & 33.9 & 42.1 \\
    LSS~\cite{philion2020lift} & ResNet-18 & C & 75.4 & 38.8 & 46.3 & 30.3 & 39.1 & 36.5 & 44.4 \\
    CVT~\cite{zhou2022crossview} & EfficientNet-B4 & C & 74.3 & 36.8 & 39.9 & 25.8 & 35.0 & 29.4 & 40.2 \\
    M\textsuperscript{2}BEV~\cite{xie2022m2bev} & ResNeXt-101 & C & 77.2 & \xm & \xm & \xm & \xm & 40.5 & \xm\\
    BEVFusion~\cite{liu2022bevfusion} & Swin-T & C & 81.7 & 54.8 & 58.4 & 47.4 & 50.7 & 46.4 & 56.6 \\
    \rowcolor{gray9} \textbf{X-Align}$_{\mathrm{view}}$ & Swin-T & C & \textbf{82.4} & \textbf{55.6} & \textbf{59.3} & \textbf{49.6} & \textbf{53.8} & \textbf{47.4} & \textbf{58.0} \\
    \midrule
    PointPillars~\cite{lang2019pointpillars} & VoxelNet & L & 72.0 & 43.1 & 53.1 & 29.7 & 27.7 & 37.5 & 43.8 \\
    CenterPoint~\cite{yin2021center} & VoxelNet & L & 75.6 & 48.4 & 57.5 & 36.5 & 31.7 & 41.9 & 48.6 \\
    \midrule
    PointPainting~\cite{vora2020pointpainting} & ResNet-101, PointPillars & C + L & 75.9 & 48.5 & 57.1 & 36.9 & 34.5 & 41.9 & 49.1\\
    MVP~\cite{yin2021multimodal} & ResNet-101, VoxelNet & C + L & 76.1 & 48.7 & 57.0 & 36.9 & 33.0 & 42.2 & 49.0\\
    BEVFusion~\cite{liu2022bevfusion} & Swin-T, VoxelNet & C + L & 85.5 & 60.5 & 67.6 & 52.0 & 57.0 & 53.7 & 62.7 \\
    \rowcolor{gray9} \textbf{X-Align}$_{\mathrm{losses}}$ & Swin-T, VoxelNet & C + L & 85.8 & 63.1 & 68.6 & 53.6 & \textbf{57.9} & 56.7 & 64.3 \\
    \rowcolor{gray8} \textbf{X-Align}$_{\mathrm{all}}$ & Swin-T, VoxelNet & C + L & \textbf{86.8} & \textbf{65.2} & \textbf{70.0} & \textbf{58.3} & 57.1 & \textbf{58.2} & \textbf{65.7}\\
    \bottomrule
    \end{tabular}
    \caption{\textbf{Quantitative evaluation on the nuScenes validation set}, in terms of single-class IoUs and the overall mIoU. We compare with existing methods from literature, where the numbers are taken from~\cite{liu2022bevfusion}, as their reproduced results are better than the ones originally reported in the papers due to a higher number of trained classes. We also provide information on the backbones and input modalities in the table. Our proposed \textbf{X-Align} outperforms all existing approaches in both single-class IoUs and the overall mIoU by a significant margin.}
    \label{tab:nuscenes_seg}
\end{table}
\begin{table}[h]
\fontsize{8pt}{9pt}\selectfont
\centering
    \setlength{\tabcolsep}{7pt}
    \renewcommand{\arraystretch}{1.2}
    \centering
    \begin{tabular}{lcc|cc}
        \toprule
        \textit{Model} & \textit{Encoder} & \textit{Modality} & \textit{mIoU} & \textit{PQ} \\
        \midrule
        PanopticBEV~\cite{gosala22panopticbev} & EffDet-D3 & C & 25.4 & 16.0 \\
        \rowcolor{gray9} 
        \textbf{X-Align$_{view}$} & EffDet-D3 & C & \textbf{27.8} & \textbf{16.9} \\
        \bottomrule
    \end{tabular}
    \caption{\textbf{Quantitative evaluation on KITTI-360} in terms of mIoU and PQ, using the camera-only modality.}
    \label{tab:kitti}
\end{table}
\subsection{Experimental Setup}
\vspace{-4pt}

\textbf{Datasets:} We evaluate performance on the large-scale nuScenes benchmark~\cite{caesar2020nuscenes}, which provides ground-truth annotations to support BEV segmentation. It contains 40,000 annotated keyframes captured by a 32-beam LiDAR scanner and six monocular cameras providing a $360^\circ$ field of view. Following the BEV map segmentation setup from~\cite{liu2022bevfusion}, we predict six semantic classes: drivable lanes, pedestrian crossings, walkways, stop lines, carparks, and lane dividers.
We further evaluate on KITTI-360~\cite{liao2022kitti}, a large-scale dataset with 83,000 annotated frames, including data collected using two fish-eye cameras and a perspective stereo camera. KITTI-360 does not provide dense ground-truth annotations for BEV segmentation. Hence, we use the BEV segmentation annotations from~\cite{gosala22panopticbev} as ground truth. These contain both static classes such as \textit{road} and \textit{sidewalk}, along with dynamic objects such as \textit{cars} and \textit{trucks}.\par
\textbf{Evaluation Metrics:} For BEV map segmentation, our primary evaluation metric is the mean Intersection Over Union (mIoU). Because some classes may overlap, we apply binary segmentation separately to each class and choose the highest IoU over different thresholds. We then take the mean over all semantic classes to produce the mIoU. This evaluation protocol was proposed in~\cite{liu2022bevfusion}. We additionally use Panoptic Quality (PQ)~\cite{kirillov2019panoptic} on KITTI-360 when evaluating panoptic BEV segmentation. 

To evaluate performance on the 3D object detection task, we use the mean Average Precision (mAP) and the nuScenes Detection Score (NDS) metrics, as defined in~\cite{caesar2020nuscenes}.

\begin{figure}[h]

    \centering
    \includegraphics[width=0.8\columnwidth]{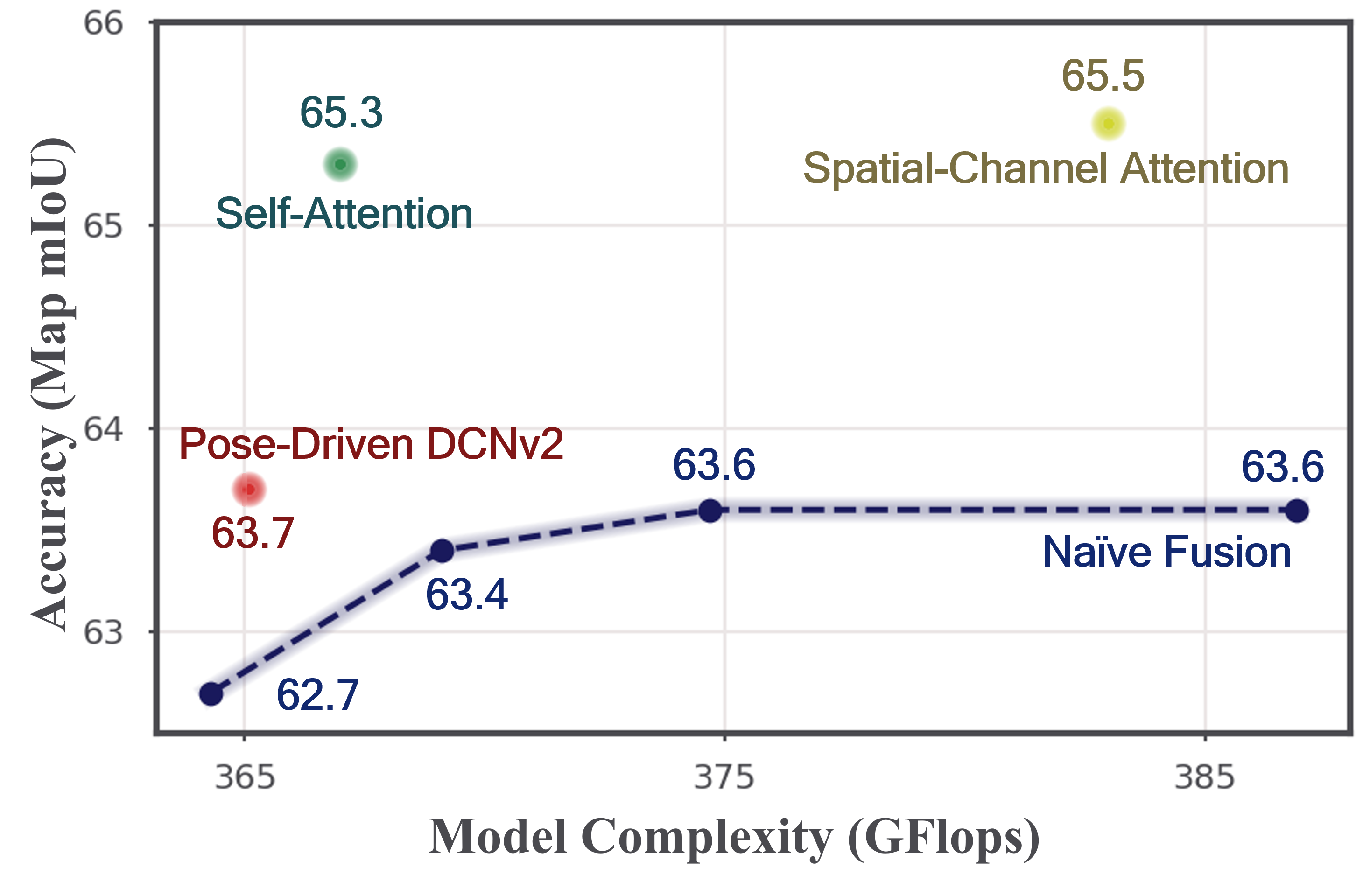}
    \caption{\textbf{Accuracy-Computation Analysis}: We compare our proposed Cross-Modal Feature Fusion (X-FF) designs with simply scaling up the fusion mechanism (adopted by existing methods~\cite{liu2022bevfusion, li2021hdmapnet}) in terms of accuracy and computation complexity.}  
    \label{fig:gmacs}
\end{figure}

\textbf{Network Architecture and Training}: For evaluation on nuScenes, we build upon BEVFusion~\cite{liu2022bevfusion} for the baseline and train our networks within mmdetection3d~\cite{mmdet3d2020}.
In the camera branch, images are downsampled to 256$\times$704 before going into a Swin-T~\cite{liu2021swin} or ConvNeXt~\cite{liu2022convnext} backbone pretrained on ImageNet~\cite{russakovsky2015imagenet}. The extracted features are fed into several FPN~\cite{lin2017feature} layers and then through a PV-to-BEV transformation based on LSS~\cite{philion2020lift} to be mapped into the BEV space. 
In the LiDAR branch, we voxelize the points with a grid size of 0.1m and use a sparse convolution backbone~\cite{yan2018second} to extract the features, which are then flattened onto the BEV space. Given the camera and LiDAR features in BEV space, we utilize our proposed X-FF mechanism from Section~\ref{sec:feature_fusion} to fuse them. We use the self-attention module in our main results, providing the best accuracy-computation trade-off (see Fig.~\ref{fig:gmacs}). The fused features are fed into a BEV encoder and FPN layers similar to those in SECOND~\cite{yan2018second} and subsequently to a segmentation head as in BEVFusion~\cite{liu2022bevfusion}. Since nuScenes does not provide ground-truth PV segmentation labels, we utilize a SOTA model pre-trained on Cityscapes to generate pseudo-labels for supervising our PV segmentation in X-SA. Specifically, we use an HRNet-w48~\cite{wang2020deep} trained with InverseForm~\cite{borse2021inverseform}.

On KITTI-360, we take the camera-only PanopticBEV~\cite{gosala22panopticbev} as the baseline, which we retrain using the code and hyperparameters released by the authors.
Then, we include our two proposed X-SA losses on top of this baseline to generate the X-Align$_{\mathrm{view}}$ results on KITTI-360.

Additional hyperparameters and training details for all experiments can be found in the Appendix.

\begin{figure*}[h]
    \centering
    \includegraphics[width=0.96\textwidth]{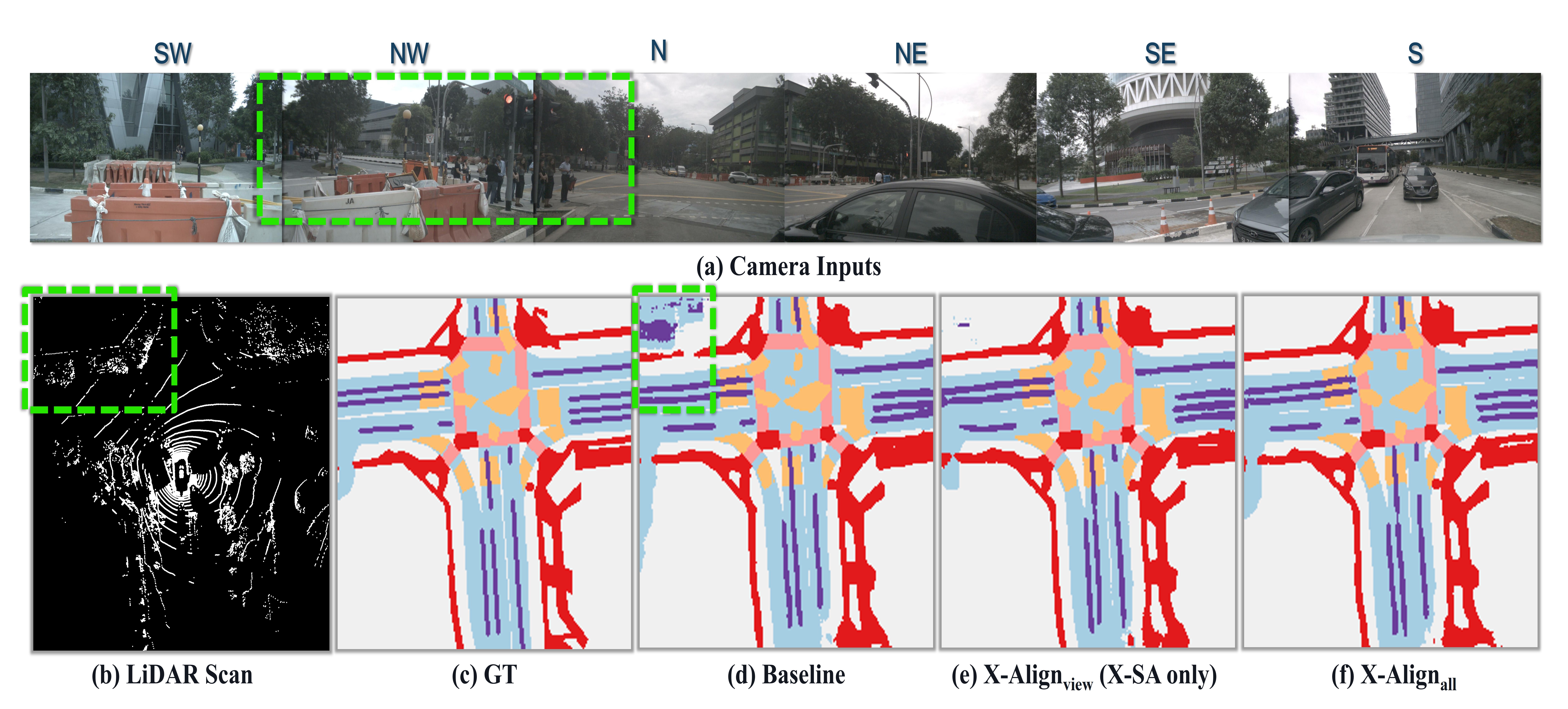}
    \caption{\textbf{Scenario 1 on nuScenes:} In this scene, the baseline wrongly predicts the building in the NW image as part of a road in the BEV segmentation output due to the inaccurate PV-to-BEV transformation. Since the building is not captured in the LiDAR scan, the LiDAR branch also cannot correct the camera projection later in the fusion. However, by utilizing our Cross-View Segmentation Alignment (X-SA), this erroneous projection can be largely rectified, as shown in (e). The remnants of this error are then completely removed when we apply our proposed alignment and fusion schemes, X-FA and X-FF, which enables proper fusion of the visual information from the camera and the geometric information from the LiDAR.}   
    \label{fig:qualitative}
\end{figure*}

\subsection{Quantitative Evaluation} 
\vspace{-4pt}


\textbf{nuScenes Camera-LiDAR Fusion:} We report segmentation results based on camera-LiDAR fusion in the bottom section of Table~\ref{tab:nuscenes_seg}. We evaluate in the region-bound of [-50m, 50m]$\times$[-50m, 50m] around the ego car following the standard evluation procedure on nuScenes~\cite{philion2020lift, zhou2022crossview, xie2022m2bev, li2022bevformer, liu2022bevfusion}. 
We compare our complete method \textbf{X-Align}$_{\mathrm{all}}$ with existing SOTA approaches on BEV segmentation and see that X-Align significantly outperforms them. Specifically, X-Align achieves a new record mIoU of \textbf{65.7\%} on nuScenes BEV segmentation and consistently improves across all the classes, thanks to the proposed novel cross-modal and cross-view alignment strategies. We used the Self-Attention block illustrated in Figure~\ref{fig:method_detail1} as our preferred X-FF strategy, as it provided the optimal trade-off in Figure~\ref{fig:gmacs}. For this strategy, the computational overhead (\textbf{0.8\%}) and increase in latency (\textbf{2\%}) is minimal as observed in Table~\ref{tab:nuscenes_seg}. We further show the performance of X-Align but without using the more advanced fusion module, \textit{i.e.}, \textbf{X-Align}$_{\mathrm{losses}}$, in the second last row of Table~\ref{tab:nuscenes_seg}. X-Align still significantly outperforms existing methods even without introducing additional computational complexity during inference. 


\textbf{nuScenes Camera-Only:} To demonstrate the efficacy of our X-SA scheme, we evaluate an instance of X-Align, \textit{i.e.}, \textbf{X-Align}$_{\mathrm{view}}$, using only the camera branch for BEV map segmentation. The results and comparisons are shown in the top part of Table~\ref{tab:nuscenes_seg}. It can be seen that \textbf{X-Align}$_{\mathrm{view}}$ considerably outperforms the existing best performance, achieving a record mIoU of \textbf{58.0\%}, surpassing the prior camera-only SOTA by \textbf{1.4} points in mIoU. This shows the benefit of X-SA, which enhances the intermediate semantic features and the PV-to-BEV transformation without incurring computational overhead at inference.\par

\textbf{KITTI-360 Camera-Only:} We present additional camera-only results on KITTI-360 in Table~\ref{tab:kitti}, for the task of panoptic BEV segmentation. In this case, we use PanopticBEV~\cite{gosala22panopticbev} as our baseline. Additionally, we include our two novel X-SA losses to train our \textbf{X-Align}$_{\mathrm{view}}$ model.\footnote{The baseline scores are obtained by training PanopticBEV using the authors' code in \href{https://github.com/robot-learning-freiburg/PanopticBEV}{https://github.com/robot-learning-freiburg/PanopticBEV.}}  It can be seen that our proposed approach improves the baseline in terms of both mIoU and PQ scores. 

Overall, our proposed X-Align consistently improves upon the existing methods across modalities and classes on both nuScenes and KITTI-360, demonstrating the efficacy of our proposed X-FF, X-FA, and X-SA components.

\begin{figure*}[h]
    \centering
    \includegraphics[width=0.98\textwidth]{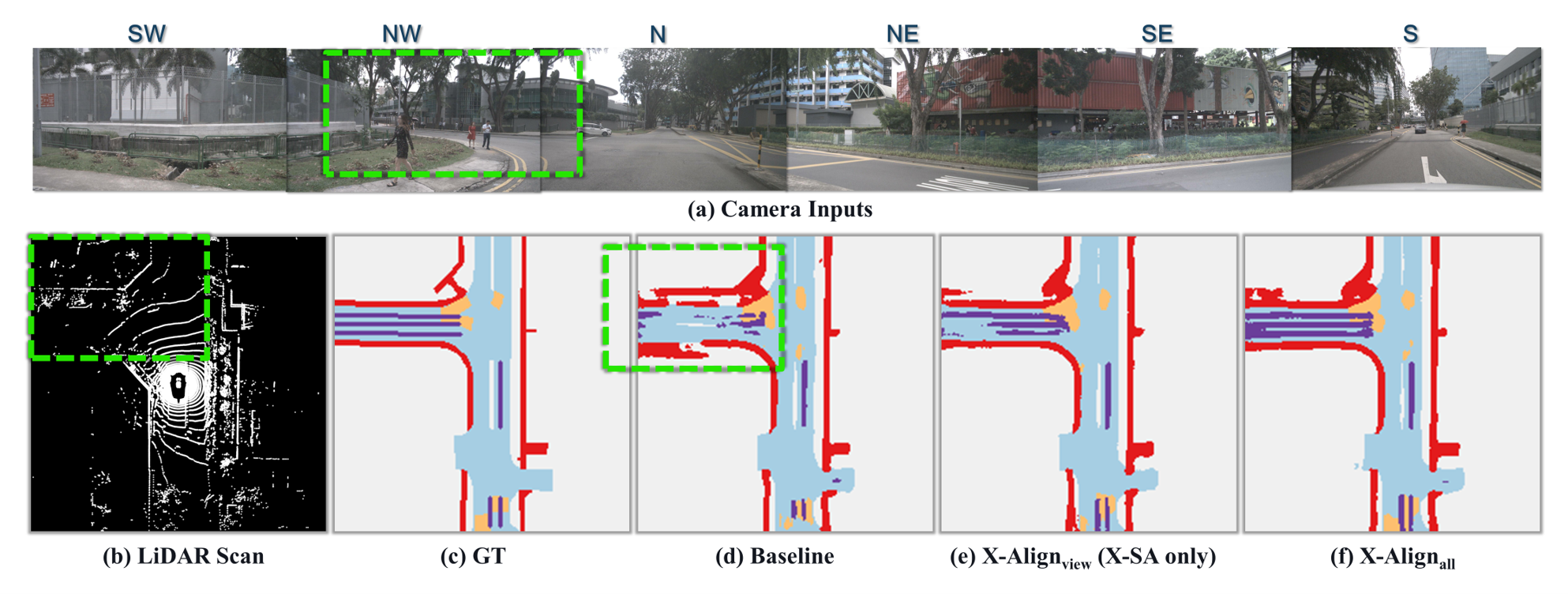}
    \caption{\textbf{Scenario 2 on nuScenes:}
    We present a scene where multiple road occlusions are apparent in the N and NW camera images, which were caused by pedestrians. The baseline model fails to fill in the gaps properly through the LiDAR-camera fusion, producing an entangled segmentation representation of the intersecting roads. Using the two X-SA losses, \textbf{X-Align}$_\text{view}$ improves the segmentation map's prediction. By adding all the components, \textbf{X-Align}$_\text{all}$ can address the occlusions and produce a more refined semantic representation of the scene.}
    \label{fig:qualitative_s1}
\end{figure*}

\begin{figure*}[h]
    \centering
    \includegraphics[width=0.98\textwidth]{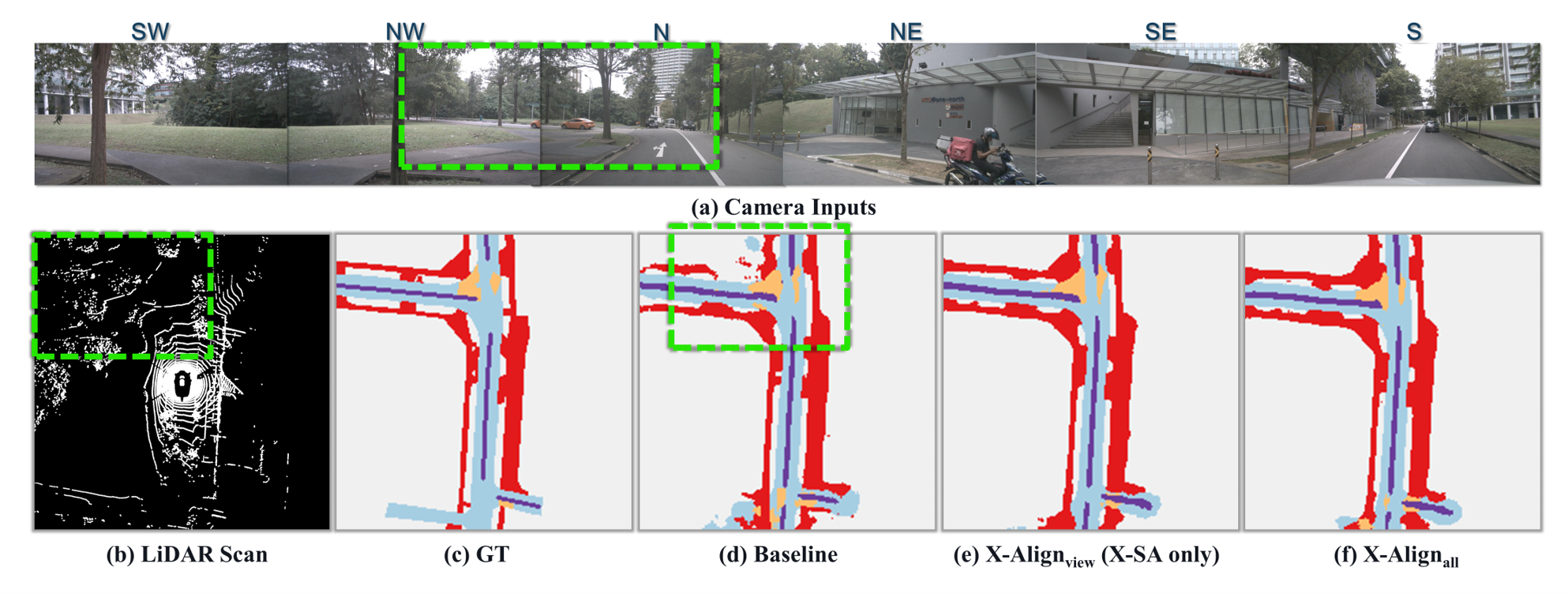}
    \caption{\textbf{Scenario 3 on nuScenes:} In this scene from nuScenes, the road boundaries in the north west direction of the vehicle appear vague in the LiDAR point cloud's green box area. However, in the N and NW camera images, they are apparent. The baseline model's boundary prediction is faulty due to misaligned camera features. Using the two X-SA losses, \textbf{X-Align}$_\text{view}$ rectifies some of the camera features. By adding all the components, \textbf{X-Align}$_\text{all}$ can predict a map where the road boundaries are properly segmented.}
    \label{fig:qualitative_s2}
\end{figure*}

\begin{figure*}[h]
    \centering
    \includegraphics[width=0.98\textwidth]{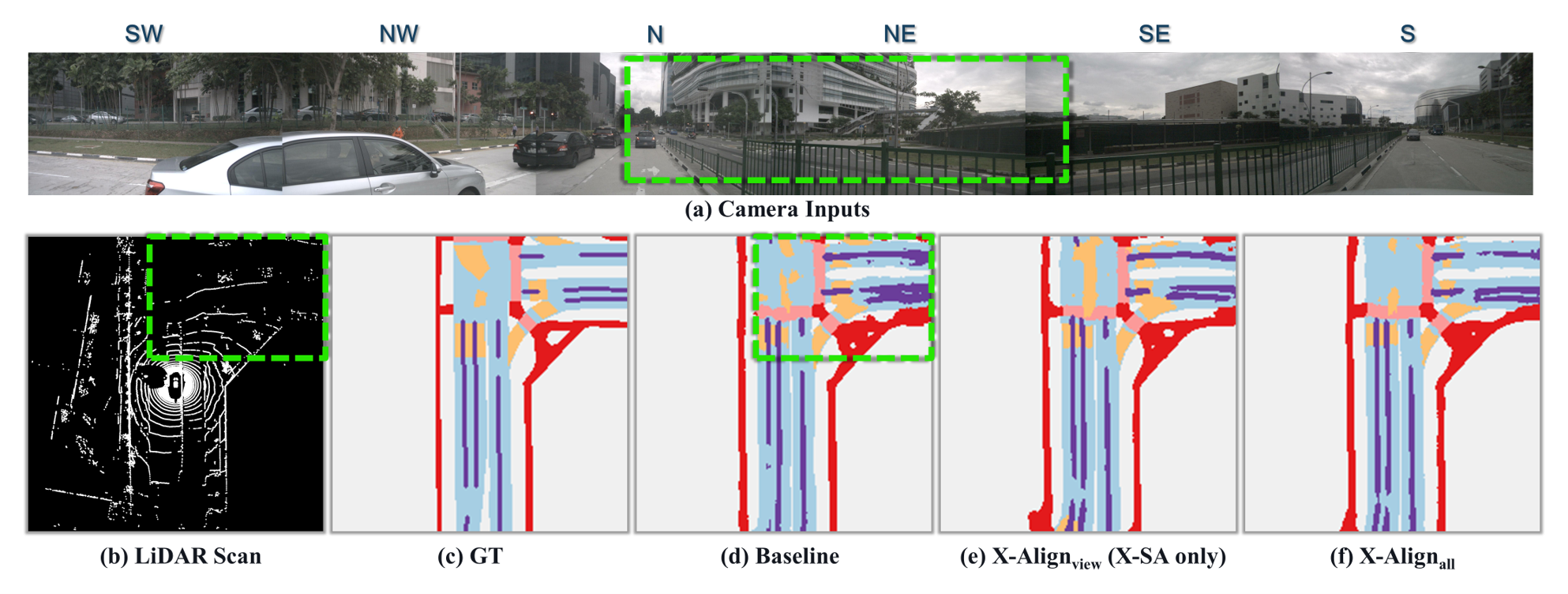}
    \caption{\textbf{Scenario 4 on nuScenes:} We present a scene where patterned occlusions like fences are evident in the N and NW camera images. Due to these repetitive occlusions, the baseline model has difficulty producing a clear segmentation map of the intersection. By using the two X-SA losses, \textbf{X-Align}$_\text{view}$ can extract more BEV-segmentation-oriented features from the images. By adding all the components, \textbf{X-Align}$_\text{all}$ can properly rectify the camera's incomplete view of the road and align it properly with the LiDAR's features, producing a clear segmentation map of the road as shown in (f).}
    \label{fig:qualitative_s3}
\end{figure*}

\begin{figure*}[h]
    \centering
    \includegraphics[width=0.98\textwidth]{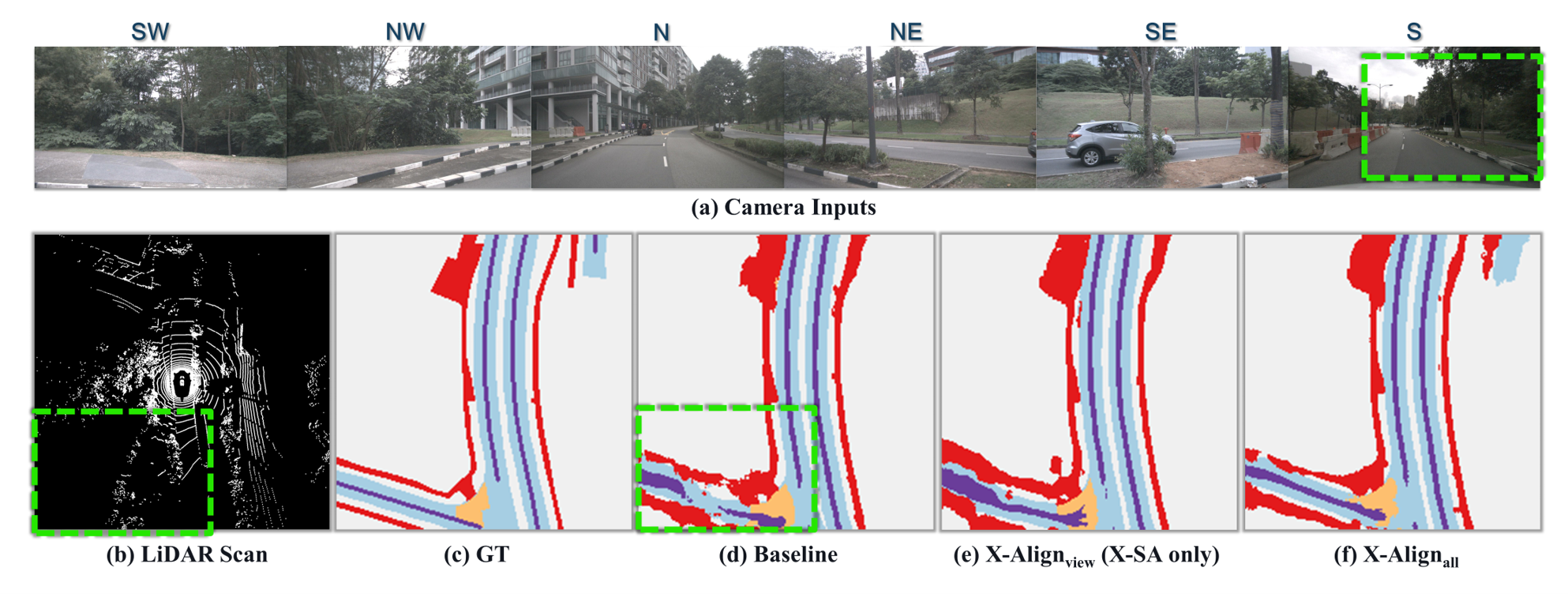}
    \caption{\textbf{Scenario 5 on nuScenes:} We present a scene from nuScenes, where the back intersection is dimly lighted with slightly occluded road boundaries, as shown in the S camera image. The LiDAR point cloud input lacks salient features capturing the road boundaries. We show the baseline's erroneous boundary prediction due to their simple concatenation-based fusion between the two modalities. We also show that by using two X-SA losses, \textbf{X-Align}$_\text{view}$ can have a more accurate prediction of the road boundaries, and by additionally using X-FF and X-FA, \textbf{X-Align}$_\text{all}$ can output an accurate segmentation map of the intersection.} 
    \label{fig:qualitative_s4}
\end{figure*}

\subsection{Accuracy-Computation Analysis}
\vspace{-4pt}
In Fig.~\ref{fig:gmacs}, we report the accuracy-computation trade-off by utilizing our different X-FF fusion strategies, including self-attention, spatial-channel attention, and pose-driven deformable convolution (DCNv2). It can be seen that when using spatial-channel attention, we achieve the highest accuracy improvement at a higher computational cost, while pose-driven DCNv2 introduces the least amount of additional cost but provides less performance gain. Using self-attention, on the other hand, provides the best trade-off between performance and complexity.

We further compare by naively scaling up the complexity of the baseline fusion, \textit{eg.}, by adding more layers and channels in the convolution blocks, shown by the blue curve. It can be seen that the baseline's performance saturates, and all our proposed fusion methods achieve better trade-offs as compared to the baseline. This again verifies that the baseline fusion using simple concatenation and convolutions does not provide the suitable capacity for the model to align and aggregate multi-modal features.
\subsection{Qualitative Results}
\vspace{-4pt}


In Figs.~\ref{fig:qualitative},~\ref{fig:qualitative_s1},~\ref{fig:qualitative_s2},~\ref{fig:qualitative_s3}, and~\ref{fig:qualitative_s4}, we present qualitative results on five sample test scenes from nuScenes, showing both LiDAR and camera inputs. We compare the BEV segmentation maps of different models, including the baseline, X-Align$_{\mathrm{view}}$ (only using the two X-SA losses), and the full X-Align, \textit{i.e.}, X-Align$_{\mathrm{all}}$.

For each of these figures, we present a scenario from the validation set in nuScenes: a) six camera inputs, b) LiDAR scan, c) ground-truth BEV segmentation map, d) baseline BEV segmentation, e) BEV segmentation using \textbf{X-Align}$_{\mathrm{view}}$, and d) BEV segmentation \textbf{X-Align}$_{\mathrm{all}}$. For each image, we observe that the baseline model prediction is highly erroneous in the region highlighted in \textcolor{green}{green}. We highlight this region of interest in the input views as well. By using the two X-SA losses, \textbf{X-Align}$_{\mathrm{view}}$ can already correct substantial errors in the baseline prediction, and the \textbf{X-Align}$_{\mathrm{all}}$ model further improves accuracy.\par

\section{Ablation Studies}
\label{sec:ablation_all}
In this section, we perform various ablation studies to evaluate each component of the X-Align framework.
\begin{table*}[h]
\fontsize{5.5pt}{6.5pt}\selectfont
    \centering
    \setlength{\tabcolsep}{3.5pt}
    \begin{tabular}{l|cccc|cccccc|ccc}
        \toprule
        \textit{Model} & 
        PV & X-FA & PV2BEV & X-FF &
        \cellcolor[HTML]{96bbce}\textit{Drivable} & 
        \cellcolor[HTML]{fb9a99}\textit{Ped. Cross.} & 
        \cellcolor[HTML]{fc4538}\textit{Walkway} & 
        \cellcolor[HTML]{fdbf6f}\textit{Stop} 
        \textit{Line} & 
        \cellcolor[HTML]{ff7f00}\textit{Carpark} & 
        \cellcolor[HTML]{ab9ac0}\textit{Divider} & 
        \cellcolor[HTML]{a5eb8d}\textit{mIoU} &
        \textit{GFlops} &
        \textit{fps} \\
        \midrule
         Baseline & \xm & \xm & \xm & \xm &  85.5 & 60.5 & 67.6 & 52.0 & 57.0 & 53.7 & 62.7 & 364.3 & 5.1 \\
         & \ch & \xm & \xm & \xm & 85.7 & 62.8 & 68.4 & 52.4 & 56.5 & 56.1 & 63.7 & 364.3 & 5.1 \\
         & \xm & \ch & \xm & \xm & 85.6 & 62.3 & 68.2 & 51.6 & 56.4 & 55.9 & 63.4 & 364.3 & 5.1 \\
         \rowcolor{gray9}
         \textbf{X-Align}$_{\mathrm{view}}$ & \ch & \xm & \ch & \xm & 85.8 & 63.1 & 68.6 & 53.2 & 57.7 & 56.4 & 64.1 & 364.3 & 5.1 \\
         \rowcolor{gray9} 
         \textbf{X-Align}$_{\mathrm{losses}}$ & \ch & \ch & \ch & \xm & 85.8 & 63.1 & 68.6 & 53.6 & 57.9 & 56.7 & 64.3 & 364.3 & 5.1 \\
        \midrule
         & \xm & \xm & \xm & \ch  &  \textbf{86.8} & 64.3 & 69.5 & 54.5 & \textbf{59.5} & 57.6 & 65.3 & 367.4 & 5.0 \\
         & \xm & \ch & \xm & \ch  & \textbf{86.8} & 65.1 & 69.8 & \textbf{60.0} & 56.5 & 58.2 & 65.4 & 367.4 & 5.0 \\ 
         & \ch & \xm & \ch & \ch  & \textbf{86.8} & 65.0 & \textbf{70.0} & 55.9 & 57.0 & 58.1 & 65.5 & 367.4 & 5.0 \\
        \rowcolor{gray8} 
        \textbf{X-Align}$_{\mathrm{all}}$ & \ch & \ch & \ch & \ch & \textbf{86.8} & \textbf{65.2} & \textbf{70.0} & 58.3 & 57.1 & \textbf{58.2} & \textbf{65.7} & 367.4 & 5.0 \\
        \bottomrule
    \end{tabular}
    \caption{\textbf{Ablation study on the proposed X-Align components.} In the top part, we show the effects of our proposed losses, \textit{i.e.}, the Cross-Modal Feature Alignment (X-FA) loss and Cross-View Segmentation Alignment (X-SA) comprised of the PV and PV2BEV segmentation losses. In the bottom part, we further show the improvements enabled by our Cross-Modal Feature Fusion (X-FF).}
    \label{tab:ablation}
\end{table*}

\subsection{Studying the effect of each component in isolation}
\label{sec:ablation}
We conduct an ablation study on the different X-Align components and summarize our results in Table~\ref{tab:ablation}. We evaluate model variants using different combinations of our proposed novel losses, including Cross-Modal Feature Alignment (X-FA) loss and our Cross-View Segmentation Alignment (X-SA) comprised of our PV and PV2BEV losses. Additionally, we investigate the effect of our Cross-Modal Feature Fusion (X-FF). 

Our X-Align$_{\mathrm{view}}$ variant, leveraging both PV and PV2BEV losses, improves the mIoU from \textbf{62.7\%} to \textbf{64.1\%} when compared to the baseline. Specifically, the PV loss alone contributes to a 1-point mIoU improvement. When using the X-FA loss, we increase the baseline mIoU from \textbf{62.7\%} to \textbf{63.4\%}. Finally, adding all the losses together, our X-Align$_{\mathrm{losses}}$ variant boosts the mIoU score to \textbf{64.3\%}, significantly surpassing the baseline's mIoU. Notably, these improvements rely on new training losses and have no computational overhead during inference.


Next, we study the effect of our X-FF module. Without any new losses, X-FF enables a 2.6-point mIoU improvement over the baseline (with a minor increase of computation, cf.~Table~\ref{tab:ablation} and Fig.~\ref{fig:gmacs}). This shows that simple concatenation is a key limitation in the baseline, which fails to fuse the camera and LiDAR features properly. Finally, using all our novel loss functions together with the X-FF module, we arrive at the full X-Align$_{\mathrm{all}}$ model, which achieves the overall best performance of \textbf{65.7\%} mIoU, significantly higher than the baseline's \textbf{ 62.7\%} mIoU. Our extensive ablation study results show that each of our proposed components in X-Align provides a meaningful contribution to improving the SOTA BEV segmentation performance.\par

\subsection{Varying Backbones}

\begin{table}[h]
\fontsize{8pt}{9pt}\selectfont
\centering
    \setlength{\tabcolsep}{7pt}
    \renewcommand{\arraystretch}{1.2}
    \centering
    \begin{tabular}{lcc|c}
        \toprule
        \textit{Model} & \textit{Encoder} & \textit{Modality} & \textit{mIoU} \\
        \midrule
        BEVFusion~\cite{liu2022bevfusion} & Swin-T & Camera, LiDAR & 62.7 \\
        \rowcolor{gray9} 
        \textbf{X-Align}$_{\mathrm{view}}$ & Swin-T & Camera, LiDAR & 64.1 \\
        \midrule
        BEVFusion & 
        ConvneXt-T & Camera, LiDAR & 62.1 \\
        \rowcolor{gray9} 
        \textbf{X-Align}$_{\mathrm{view}}$ & ConvneXt-T & Camera, LiDAR & 63.8 \\
        \midrule
        BEVFusion &  
        ConvneXt-S & Camera, LiDAR & 63.9 \\
        \rowcolor{gray9} 
        \textbf{X-Align}$_{\mathrm{view}}$ & ConvneXt-S & Camera, LiDAR & \textbf{64.7}
        \\
        \bottomrule
    \end{tabular}
    \caption{\textbf{Quantitative evaluation on nuScenes} in terms of mIoU, varying the camera encoder.}
    \label{tab:backbones}
\end{table}
To verify the generalizability of our X-Align method, we present results for three different encoders, \textit{i.e.}, SWin-T, ConvneXt-T, and ConvneXt-S, in Table~\ref{tab:backbones}. For this ablation study, we choose our \textbf{X-Align}$_{\mathrm{view}}$ variant because this shows that a given model can be improved without adding additional computational complexity during inference. Compared to the current state-of-the-art result from BEVFusion~\cite{liu2022bevfusion}, we can improve their result from 62.7 to 64.1 in terms of mIoU. Similar improvements can be observed for our additionally investigated backbones ConvneXt-T and ConvneXt-S. These results show that our method generalizes well to different backbones without hyperparameter tuning.
\subsection{Adaptability to Noise}

\begin{table}[h]
\fontsize{8pt}{9pt}\selectfont
\centering
    \setlength{\tabcolsep}{5pt}
    \renewcommand{\arraystretch}{1.2}
    \centering
    \begin{tabular}{lcc|cccc}
        \toprule
        \textit{Model} & \textit{Encoder} & \textit{Modality} & 
        \textit{$\sigma=0.0$} &
        \textit{$\sigma=0.05$} &
        \textit{$\sigma=0.075$} &
        \textit{$\sigma=0.1$} \\
        \midrule
        BEVFusion~\cite{liu2022bevfusion} & Swin-T & Camera &  56.6 & 52.7 & 47.2 & 40.7\\
        \rowcolor{gray9} 
        \textbf{X-Align}$_{\mathrm{view}}$ & Swin-T & Camera & \textbf{58.0} & \textbf{55.2} & \textbf{51.4} & \textbf{45.6} \\
        \midrule
        BEVFusion~\cite{liu2022bevfusion} & Swin-T & Camera, Lidar &  62.7 & 59.2 & 54.8 & 48.9\\
        \rowcolor{gray9} 
        \textbf{X-Align}$_{\mathrm{view}}$ & Swin-T & Camera, Lidar & 64.1 & 62.1 & 58.3 & 54.1 \\
        \rowcolor{gray8} 
        \textbf{X-Align}$_{\mathrm{all}}$ & Swin-T & Camera, Lidar & \textbf{65.7} & \textbf{64.3} & \textbf{62.6} & \textbf{59.6} \\
        
        \bottomrule
    \end{tabular}
    \caption{\textbf{Quantitative evaluation on nuScenes}, adding gaussian noise to the input images. We observe that X-Align methods provide more consistent results as they are lesser variant to noise.}
    \label{tab:noise}
\end{table}

Another interesting property of deep neural network-based methods is their susceptibility to input perturbations because it gives insights into their robustness in real environments. Therefore, we add Gaussian noise to the input camera images and observe how the performance degrades in Table~\ref{tab:noise}.  The noise is zero-mean normalized, and $\sigma$ in Table~\ref{tab:noise} is the standard deviation of the Gaussian noise. From the table, we can deduce two main observations: First, the performance of X-Align methods under input perturbations is relatively stable compared to the baseline. We attribute this to the fact that the input camera noise will make the extracted camera feature less reliable. Due to our robust attention-based cross-modal feature fusion (X-FF), the method can still correct the camera features using LiDAR predictions. Our second interesting observation concerns the comparison to the baseline BEVFusion. We observe that our performance loss at a higher noise-variance is much smaller than for BEVFusion, e.g., the performance of our method \textbf{X-Align}$_{\mathrm{all}}$ at $\sigma=0.1$ drops from 65.7 to 59.6. In contrast, BEVFusion drops from 62.7 to 48.9. This shows superior properties of our method in terms of robustness.\par
\subsection{Adverse Weather Conditions}

\begin{table}[h]
\fontsize{8pt}{9pt}\selectfont
\centering
    \setlength{\tabcolsep}{4pt}
    \renewcommand{\arraystretch}{1.2}
    \centering
    \begin{tabular}{lcc|ccc}
        \toprule
        \textit{Model} & \textit{Encoder} & \textit{Modality} & 
        \textit{nuScenes-Night} &
        \textit{nuScenes-Rain} &
        \textit{nuScenes} \\
        \midrule
        BEVFusion~\cite{liu2022bevfusion} & Swin-T & Camera &  30.8 & 50.5 & 56.6 \\
        \rowcolor{gray9} 
        \textbf{X-Align$_{view}$} & Swin-T & Camera & \textbf{33.1} & \textbf{51.1} & \textbf{58.0}  \\
        \midrule
        BEVFusion~\cite{liu2022bevfusion} & Swin-T & Camera, LiDAR &  43.6 & 55.9 & 62.7 \\
        \rowcolor{gray9} 
        \textbf{X-Align}$_{\mathrm{view}}$ & Swin-T & Camera, LiDAR & 44.5 & 56.3 & 64.3  \\
        \rowcolor{gray8} 
        \textbf{X-Align}$_{\mathrm{all}}$ & Swin-T & Camera, LiDAR & \textbf{46.1} & \textbf{57.8} & \textbf{65.7}  \\
        \bottomrule
    \end{tabular}
    \caption{\textbf{Quantitative evaluation on nuScenes-rainy and nuScenes-night}. We observe that X-Align provides consistent improvements in adverse weather conditions.}
    \label{tab:splits}
\end{table}

To further investigate our method's robustness, we show results of how our method performs in adverse weather conditions in Table~\ref{tab:splits}. For this experiment, we use the metadata provided by nuScenes, which indicates the weather and light conditions of the recorded scene. With this information, we filter the validation set to obtain two splits containing all images recorded at night and all images recorded during rainy conditions. Our results show consistent improvement in these adverse conditions by \textbf{X-Align}$_{\mathrm{view}}$ as well as \textbf{X-Align}$_{\mathrm{all}}$. This shows that our method improves on comparably easy samples and difficult ones, which is an important property for future deployment of our method.

\subsection{Extending the results to Camera and Radar Object Detection}

\begin{table}[h]
\fontsize{8pt}{9pt}\selectfont
\centering
    \setlength{\tabcolsep}{7pt}
    \renewcommand{\arraystretch}{1.2}
    \centering
    \begin{tabular}{lc|cc}
        \toprule
        \textit{Model} & \textit{Modality} & \textit{mAP} & \textit{NDS} \\
        \midrule
        BEVFusion~\cite{liu2022bevfusion} & Camera, Radar & 38.9 & 51.0\\
        \rowcolor{gray9} 
        \textbf{X-FF$_{pose}$} & Camera, Radar & \textbf{39.7} & \textbf{52.1} \\
        \bottomrule
    \end{tabular}
    \caption{\textbf{Results on nuScenes detection.} The pose-driven deformable convolution X-FF strategy can help improve Camera and Radar fusion compared to vanilla BEVFusion.}
    \label{tab:nuscenes_det}
\end{table}

To investigate if our proposed X-FF strategies can extend to other tasks beyond segmentation, we show results on Camera and Radar Fusion for 3D Object Detetion(3DOD). For this experiment, we first train a baseline model using on Camera and Radar inputs and fusing them using the vanilla convolution based fusion strategy from BEVFusion. We follow the baseline hyperparameters from BEVFusion for their 3DOD variant, and replace the LiDAR inputs with Radar inputs. We compare this baseline with our pose-driven deformable convolution, \textbf{X-FF}$_{\mathrm{pose}}$ described in Section~\ref{sec:feature_fusion}. We observe an improvement of \textbf{1.1} points on NDS and \textbf{0.8} points on mAP.    
\section{Conclusions}
In this paper, we proposed a novel framework, X-Align, which addresses cross-view and cross-modal alignment in BEV segmentation. It enhances the alignment of unimodal features to aid feature fusion and the alignment between perspective view and bird's-eye-view representations. Our experiments show that X-Align improves performance on nuScenes and KITTI-360 datasets, in particular outperforming previous SOTA by 3 mIoU points on nuScenes. We also verified the effectiveness and robustness of the X-Align components via extensive ablation studies.

\section{Limitations and Future work}
One of the limitations of our X-SA module is the dependence on accurate pseudo-labels when semantic ground truth isn't available. This was the case in nuScenes, for which we used an expert segmentation model to pseudo-label the perspective view images. We plan to tackle the dependence on accurate perspective view labels as part of the Future work. We also plan to further study the generalization capability of X-Align to perform more tasks in multi-modal perception such as Depth Estimation, 3D Object detection and tracking.

\begin{appendices}
\section{Training Details and Hyperparameter Analysis}

In this subsection, we provide the hyper-parameters and training details for all our experiments in the paper. All our models are trained using Pytorch~\cite{paszke2017automatic} on 4 Nvidia Tesla A100 GPUs.\par

\textbf{nuScenes experiments:} To reproduce the BEVFusion~\cite{liu2022bevfusion} baseline model, we adapt the code provided by their authors\footnote{\href{https://github.com/mit-han-lab/bevfusion}{https://github.com/mit-han-lab/bevfusion}} into the mmdetection3d~\cite{mmdet3d2020} framework. This is because they do not provide training code in the current version. In the camera pipeline, the images are downsampled into a size of $256\times704$ before being passed through a Swin-T~\cite{liu2021swin} or ConvNext~\cite{liu2022convnext} backbone pretrained on ImageNet~\cite{russakovsky2015imagenet}. The intermediate features extracted from the camera are passed through a set of FPN~\cite{lin2017feature} layers to retain the salient low-level encoder features. These are then passed to View-Transformers based on LSS~\cite{philion2020lift}. The baseline is trained using their reported hyper-parameters~\cite{liu2022bevfusion}, with a learning schedule of 20 epochs and a cyclic learning rate, starting for $1e^{-4}$ and performing a single cycle with target ratios $\{10, 1e^{-4}\}$ and a step of $0.4$. For our experiments, we implement our proposed blocks:
\begin{itemize}
\item
\textbf{X-FF Modules}: For the X-FF modules described in Section~3.3 of the paper, we switch the naive convolutional fuser in the baseline model. To implement the Self-Attention fuser, we tokenize the inputs into patches of size $3\times3$ and a stride of 2. This step is followed by a Multi-Head-Self-Attention block as described in~\cite{vaswani2017attention}, containing 8 heads and an embedding dimension of 256. To implement the Spatial-Channel Attention module, we use the Split-depth Transpose Attention (SDTA) block from EdgeNext~\cite{maaz2022edgenext}.We use an embedding dimension of 256, 2 scales, and 8 heads. We also use DropPath~\cite{larsson2016fractalnet} of $0.1$. A deconvolution block follows the SDTA fusion module to output 256 channels and the fused feature resolution. Finally, to implement the pose-driven deformable convolutional operation, we pass the input pose through a 2-layer MLP network, whose output can is interpolated to the feature size. This pose channel is concatenated with the two input modalities, then passes through a convolutional block to obtain $18$ offset channels, input into a DCNv2~\cite{zhu2019deformable} block of kernel size $3\times3$ The output of the DCNv2 block is our fused feature dimension.
\item
\textbf{X-FA Loss}: As explained in Section~3.4 of the paper, we use cosine similarity as a loss function to model the X-FA similarity. Using a sparse hyper-parameter search, we set the best value of $\gamma_{2}$ in Equation~(2) of the main paper as $-0.002$. We reduce its negative value because we want to increase the cosine similarity.
\item
\textbf{X-SA Module}: As explained in Section~3.5 of the paper, we utilize a perspective decoder to generate 2D perspective view semantic labels. The decoder has two Convolution-BN2D-ReLU sequences with a hidden dimension of $256$ channels, generating semantic probabilities at the output. As nuScenes~\cite{caesar2020nuscenes} does not contain ground truth segmentation labels; we make use of a state-of-the-art (SOTA) model pre-trained on Cityscapes to generate pseudo labels $\bm{y}^{\mathrm{PV}}$. This is an HRNet-w48~\cite{wang2020deep} checkpoint, trained with the InverseForm~\cite{borse2021inverseform} loss, as made public by the authors in their codebase.\footnote{\href{https://github.com/Qualcomm-AI-research/InverseForm}{https://github.com/Qualcomm-AI-research/InverseForm}} We pick $\gamma_{3}$ in Equation~(2) of the main paper as $0.1$ following a sparse hyper-parameter search. Once we have the 2D perspective features, we share parameters with the LSS Transform from earlier to splat the probability maps to BEV space. We use convolution blocks for channel aggregation at the input and output and interpolate to equalize dimensions. The output is then supervised with BEV GT as shown in Figure~\ref{fig:method_detail1} of the paper. The loss weight $\gamma_4$ is tuned to $0.1$ by observing the total loss value and setting the weight to produce $\sim\! 20\%$ of the total loss. 
\end{itemize}

\textbf{KITTI360 experiments:} To reproduce the PanopticBEV~\cite{gosala22panopticbev} scores, we use the code made public by the authors and their default settings.\footnote{\href{https://github.com/robot-learning-freiburg/PanopticBEV}{https://github.com/robot-learning-freiburg/PanopticBEV}} The model consists of an EfficientDet-d3~\cite{tan2020efficientdet} camera encoder, which encodes multi-resolution features. Next, a multi-scale transformer converts the perspective view to BEV features. This is followed by decoders for both instance and semantics. Finally, their outputs are combined to generate panoptic predictions. The baseline uploaded by the authors reaches a lower value than what was obtained in their paper.  For our experiments with the \textbf{X-SA module}, we added a 2D Perspective decoder to the multi-scale encoded features to predict semantic labels in the perspective view. As KITTI360 contains 2d semantic labels, we use them to supervise this decoder. Then, we reuse the view transformer to map predictions to BEV space. We supervise our X-SA branch with loss weights $\gamma_3$ set at $0.5$ and $\gamma_4$ set at $0.1$, adding them to the total loss of the PanopticBEV baseline.

\end{appendices}

\bibliography{bib/bibpaper}

\end{document}